\documentclass[11pt,twocolumn]{article}

\usepackage{graphicx}
\usepackage{subfig}
\usepackage{subfloat}
\usepackage[ruled,english,onelanguage,boxed,linesnumbered]{algorithm2e}
\usepackage{multirow}
\newcommand{\nonl}{\renewcommand{\nl}{\let\nl\oldnl}}
\usepackage{booktabs}
\usepackage{multicol}
\usepackage{caption}
\usepackage{amssymb}
\usepackage{amsmath}
\usepackage{nccmath}
\usepackage{hyperref}
\usepackage{mathptmx} 

\begin{document}

\title{\bf{Telepresence System based on Simulated Holographic Display}}

\author{Diana-Margarita C\'ordova-Esparza$^{1}$ \and Juan R. Terven$^{2}$ \and Hugo Jim\'enez-Hern\'andez$^{3}$ \and
Ana Herrera-Navarro$^{1}$ \and Alberto V\'azquez-Cervantes$^{3}$ \and Juan-M. Garc\'ia-Huerta$^{3}$ \\
\small{$^{1}$  UAQ, Universidad Aut\'onoma de Quer\'etaro \quad $^{2}$ AiFi Inc. \quad $^{3}$ CIDESI} \\
}
\date{}
\maketitle

%%%%%%%%%%%%%%%%%%%%%%%%%%%%%%%%%%%%%%%%%%%%%%%%%%%%%%%%%%%%%%%%%%%%%%%%%%%%%
%%%%%%%%%%%%%%%%%%%%%%%%%%%%%%%%%%%%%%%%%%%%%%%%%%%%%%%%%%%%%%%%%%%%%%%%%%%%%
\begin{abstract}
We present a telepresence system based on a custom-made simulated holographic display that produces a full 3D model of the remote participants using commodity depth sensors. Our display is composed of a video projector and a quadrangular pyramid made of acrylic, that allows the user to experience an omnidirectional visualization of a remote person without the need for head-mounted displays. To obtain a precise representation of the participants, we fuse together multiple views extracted using a deep background subtraction method. Our system represents an attempt to democratize high-fidelity 3D telepresence using off-the-shelf components. 
 
\end{abstract}

%%%%%%%%%%%%%%%%%%%%%%%%%%%%%%%%%%%%%%%%%%%%%%%%%%%%%%%%%%%%%%%%%%%%%%%%%%%%%
%%%%%%%%%%%%%%%%%%%%%%%%%%%%%%%%%%%%%%%%%%%%%%%%%%%%%%%%%%%%%%%%%%%%%%%%%%%%%
\section{Introduction}
\label{intro}

Telepresence is the process of reproducing the visual, auditory, or perceptual information in a remote location. With this technology, business transportation costs, which are estimated to be 1.3 trillion dollars in 2016 and are predicted to rise to 1.6 trillion in 2020 in the US alone~\cite{statistaTravel}, can be reduced significantly.

Arthur C. Clarke 1974's prediction about computers have become a reality~\cite{wikiClarke}:
\emph{``They will make it possible to live anywhere we like. Any businessman, any executive, could live almost anywhere on Earth and still do his business through a device like this``}. Although this is true, thanks to the Internet, current teleconference technology is still far apart from the actual feeling of \emph{physical co-presence}. Applications such as Skype, FaceTime, and GoToMeeting are limited in the sense that they provide only a 2D view of the participants displayed on a flat screen. One solution to this limitation is the use of holographic displays.

A Holographic display is a technology that performs reconstruction of light wavefronts by using the diffraction of coherent light sources~\cite{dreshaj2015holosuite}. This kind of displays can create images with 3D optical effects without the need for additional devices such as glasses or head-mounted displays \cite{gohane20143d}. However, building true holographic displays is costly and requires specialized hardware. For these reasons, there have been many attempts to create simulated holographic displays~\cite{bimber2005spatial,tiro2015possibility,dalvi2015,yoo2014study,antonio2013projection}. This refers to using more conventional 3D displays that use stereoscopic vision and motion parallax reprojection to approximate visual cues provided inherently in holographic images~\cite{dreshaj2015holosuite}.

\begin{figure*}
\centering
  \includegraphics[width=0.6\textwidth]{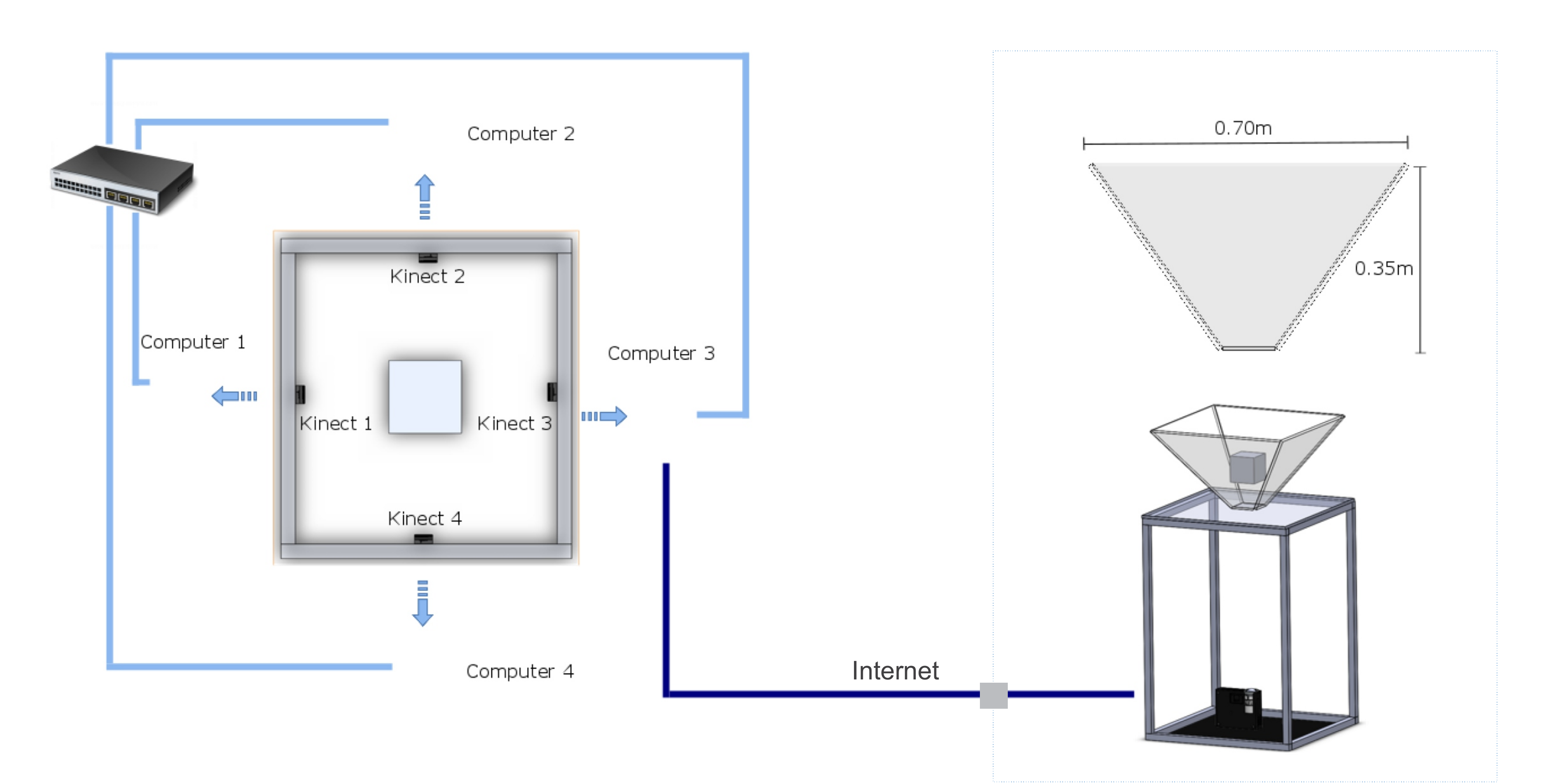}
\caption{Shows a schematic overview of the experimental setup for the proposed system, which consists of four Kinect V2 cameras, and the holographic display composed of commodity light projector and an acrylic square pyramid.}
\label{fig:1}       % Give a unique label
\end{figure*}

In this work, we introduce a real-time, full 3D telepresence system, which uses an array of depth and color cameras (RGB-D) and simulated holographic projection. We extend our existing multi-camera system~\cite{cordova2017multiple} with the ability to precisely segment the foreground objects and project them onto a custom-made fake-holographic display using a commodity projector (Figure \ref{fig:1}).

In contrast to traditional telepresence systems, the virtual representation of the remote participant is projected onto an inverted pyramid to simulate a holographic effect. The virtual participant is obtained from four RGB-D sensors producing a 3D image through data fusion and reconstruction. Our system does not require users to wear any display or tracking equipment, nor a special background. We extract the person or objects using a deep foreground segmentation method that precisely segments the person or objects of interest and at the same time reduces the amount of data that needs to be transferred to the remote location where it is rendered. 

Figure \ref{fig:1} shows a schematic overview of our system, which consists of four Kinect V2 cameras placed at 2 meters high with a viewpoint change of approximately $90^{\circ}$, a light projector, and an acrylic square pyramid as visualization platform.

Our main contribution is a novel end-to-end real-time telepresence system capable of rendering people and objects in full 3D on a simulated holographic display.

The remainder of the paper is organized as follows. In section 2, we review related literature. We describe the methodology in section 3. In section 4, we evaluate the performance of the system. We conclude in section 5.

%.....................................
% IMAGE: Schematic overview
%.....................................
%.....................................

%%%%%%%%%%%%%%%%%%%%%%%%%%%%%%%%%%%%%%%%%%%%%%%%%%%%%%%%%%%%%%%%%%%%%%%%%%%%%
%%%%%%%%%%%%%%%%%%%%%%%%%%%%%%%%%%%%%%%%%%%%%%%%%%%%%%%%%%%%%%%%%%%%%%%%%%%%%
\section{Previous work}

One of the earliest works in telepresence is the one from Towels et al.~\cite{towles20023d} which enabled end-to-end 3D telepresence with an interaction between participants. They employed an array of cameras and a pointing device to control and manipulate a shared 3D object. The image was displayed on a stereoscopic display with head-tracking. They performed 3D reconstruction to generate a point cloud of the users and transmit it over the Internet at two fps.

More recently, with the availability of consumer depth cameras along with color cameras (RGB-D), there was an exponential emergence of 3D telepresence systems. Noticeable examples are the work of Maimone et al.~\cite{Maimone:2011,Maimone2012} with a dynamic telepresence system composed of multiple Kinect sensors. Beck et al.~\cite{Beck2013} with the introduction of an immersive telepresence system that allows distributed groups of users to meet in a shared virtual 3D world, and Room2Room~\cite{pejsa2016room2room} with a life-size telepresence system based on projected augmented reality. This system is capable of understanding the structure of the environment and projecting the remote participant onto physically plausible locations.

Along with depth cameras that are used to sense the environment, head mounted displays (HMD) have been the desired choice for AR/VR visualization. Maimone et al.~\cite{maimone2013general} presented a proof-of-concept of a general purpose telepresence system using optical see-through displays. Xinzhong et al.~\cite{Xinzhong} proposed an immersive telepresence system by employing a single RGB-D and an HMD. Lee et al.~\cite{lee2016hologram,lee2017mixed} describe a telepresence platform where a user wearing an HMD can interact with remote participants that can experience the user's emotions through a small holographic display, and finally, Microsoft's Holoportation~\cite{orts2016holoportation} represents the first high-quality real-time telepresence system in AR/VR devices. They used multi-view active stereo along with sophisticated spatial audio techniques to sense the environment. The quality of immersion is unprecedented; however, the amount of high-end hardware and high-bandwidth requirements makes this system hard to reproduce. 

Similar to these works, our system uses multiple RGB-D cameras to sense the environment. However, we display the remote participant on a set of projection screens to provide a  $360^{\circ}$ simulated holographic effect.

%% Holographic telepresence
Regarding true holographic telepresence, Blanche et al~\cite{blanche2010holographic} developed the first example of this technology using a photorefractive polymer material to demonstrate a holographic display. They used 16 cameras taking pictures every second and their system can refresh images every two seconds. More recently, Dreshaj~\cite{dreshaj2015holosuite} introduced \emph{Holosuite}, an implementation of an end-to-end 3D telepresence operating on two remote PCs via the internet. It can render visual output to the holographic displays Mark II and Mark IV, as well as on commercial 3D displays such as the zSpace~\cite{noor2015potential} with motion parallax reprojection. Holosuite uses RGB-D cameras to sense the environment and allows seamless collaboration between the participants in real-time. 

In summary, while previous works address many challenges presented in telepresence systems, and the most impressive results require high-end hardware and high bandwidth, our system can render a  $360^{\circ}$ volumetric telepresence system on simulated holographic display made of off-the-shelf components without the need of obtrusive wearable devices.

%%%%%%%%%%%%%%%%%%%%%%%%%%%%%%%%%%%%%%%%%%%%%%%%%%%%%%%%%%%%%%%%%%%%%%%%%%%%%
%%%%%%%%%%%%%%%%%%%%%%%%%%%%%%%%%%%%%%%%%%%%%%%%%%%%%%%%%%%%%%%%%%%%%%%%%%%%%
\section{Methodology}\label{sec:system_comp}
\begin{figure}[!ht]
\centering
\includegraphics[scale=0.25]{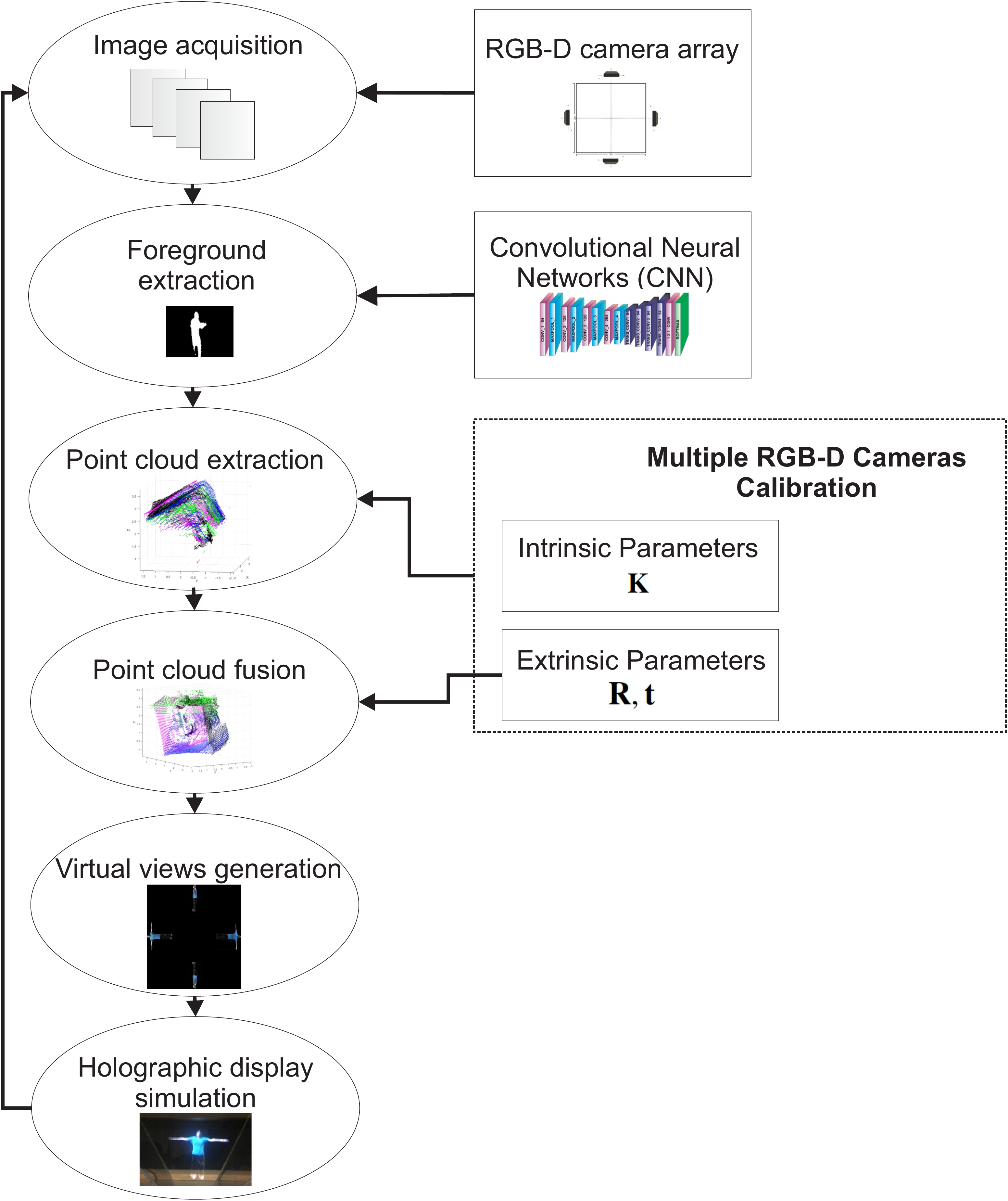}
\caption{Method followed to acquire, segment, reconstruct, fuse, and visualize 3D objects on the holographic display.}
\label{fig:methodo}
\end{figure}

In this section, we describe the procedure followed to implement our telepresence system based on simulated holographic projection. 
Figure \ref{fig:methodo} shows the steps followed to acquire, reconstruct, and visualize remote people using our 3D display.
We start by describing the camera calibration approach, followed by the foreground extraction method and data fusion. 

%.....................................
% IMAGE: Method
%.....................................
%.....................................

%%%%%%%%%%%%%%%%%%%%%%%%%%%%%%%%%%%%%%%%%%%%%%%%%%%%%%%%%%%%%%%%%%%%%%%%%%%%%
%%%%%%%%%%%%%%%%%%%%%%%%%%%%%%%%%%%%%%%%%%%%%%%%%%%%%%%%%%%%%%%%%%%%%%%%%%%%%
\subsection{Multiple RGB-D cameras calibration}
\label{sec:CamCalib}

For calibration, we followed the method from~\cite{cordova2017multiple} and it is briefly described below. Note that we assume that the cameras are fixed in the scene so that the calibration procedure is done only once.

The first step of camera calibration is image acquisition. Each RGB-D camera is connected to a single computer, and the whole system communicates through a wireless network. However, all the processing is performed on the main computer that receives and stores the calibration data. We use a 1D calibration pattern composed of a $60$ cm wand with three collinear points as shown in~\cite{cordova2017multiple}.

The color images have a resolution of $1920\times 1080$ pixels, and the depth images have a resolution of $512\times 424$ pixels. 

Once we acquire the data, we perform an initial estimation of the extrinsic parameters of the cameras on a global reference in 3D coordinate space. In our experiment, there are eight cameras in total; one depth and one color camera for each Kinect sensor.

Setting the first RGB-D camera ($i=1$) as the reference, finding the pose of the cameras involves obtaining the best rotations ${\bf{R}}_{i}$ and translations ${\bf{t}}_{i}$ that align the points from the RGB-D cameras (${\bf{M}}_{i}=[{\bf{A}}_{i}' {\bf{B}}_{i}' {\bf{C}}_{i}']$, $i \in \{2,3,4\}$) to the points in the reference RGB-D camera ($\bf{M}_1$). We wish to solve for ${\bf{R}}_{i}$ and ${\bf{t}}_{i}$ such that

\begin{equation}\label{matM}
{\bf{M}}_{1} = {\bf{R}}_{i} \times {\bf{M}}_{i} + {\bf{t}}_{i}
\end{equation} 

where ${\bf{R}}_{i}$ and ${\bf{t}}_{i}$ are the rotations and translations applied to each set of points ${\bf{M}}_{i}$, $i \in \{2,3,4\}$ to align them with the reference ${\bf{M}}_{1}$.

Once we estimate the rigid transformations that align the cameras with the reference, we applied these transformations to the point clouds $\mathbf{PC}_i$, $i\in \{2,3,4\}$  to align the 3D points from all the cameras into a single coordinate frame. Then we apply  Iterative Closest Point (ICP) on each aligned point cloud with the reference to refine the alignment.

Using the point cloud alignments, the next step is to gather multiple points for calibration. We do this by finding matches on the point clouds using nearest neighbors search. Next, we estimate the intrinsic parameters using the 3D points and the 2D projections on the image plane assuming a pinhole model for the cameras as described in~\cite{SimonPrince}.

The final calibration results are obtained through a non-linear minimization using the Levenberg-Marquard algorithm~\cite{levenberg1944method,marquardt1963algorithm} including radial and tangential distortion coefficients~\cite{brown1971lens,duane1971close}. To correct this distortion, we apply the classic model described in~\cite{brown1971lens}. \\

%%%%%%%%%%%%%%%%%%%%%%%%%%%%%%%%%%%%%%%%%%%%%%%%%%%%%%%%%%%%%%%%%%%%%%%%%%%%%
%%%%%%%%%%%%%%%%%%%%%%%%%%%%%%%%%%%%%%%%%%%%%%%%%%%%%%%%%%%%%%%%%%%%%%%%%%%%%
\subsection{Foreground Extraction}
\begin{figure*}[!ht] 
\centering
\includegraphics[scale=0.35]{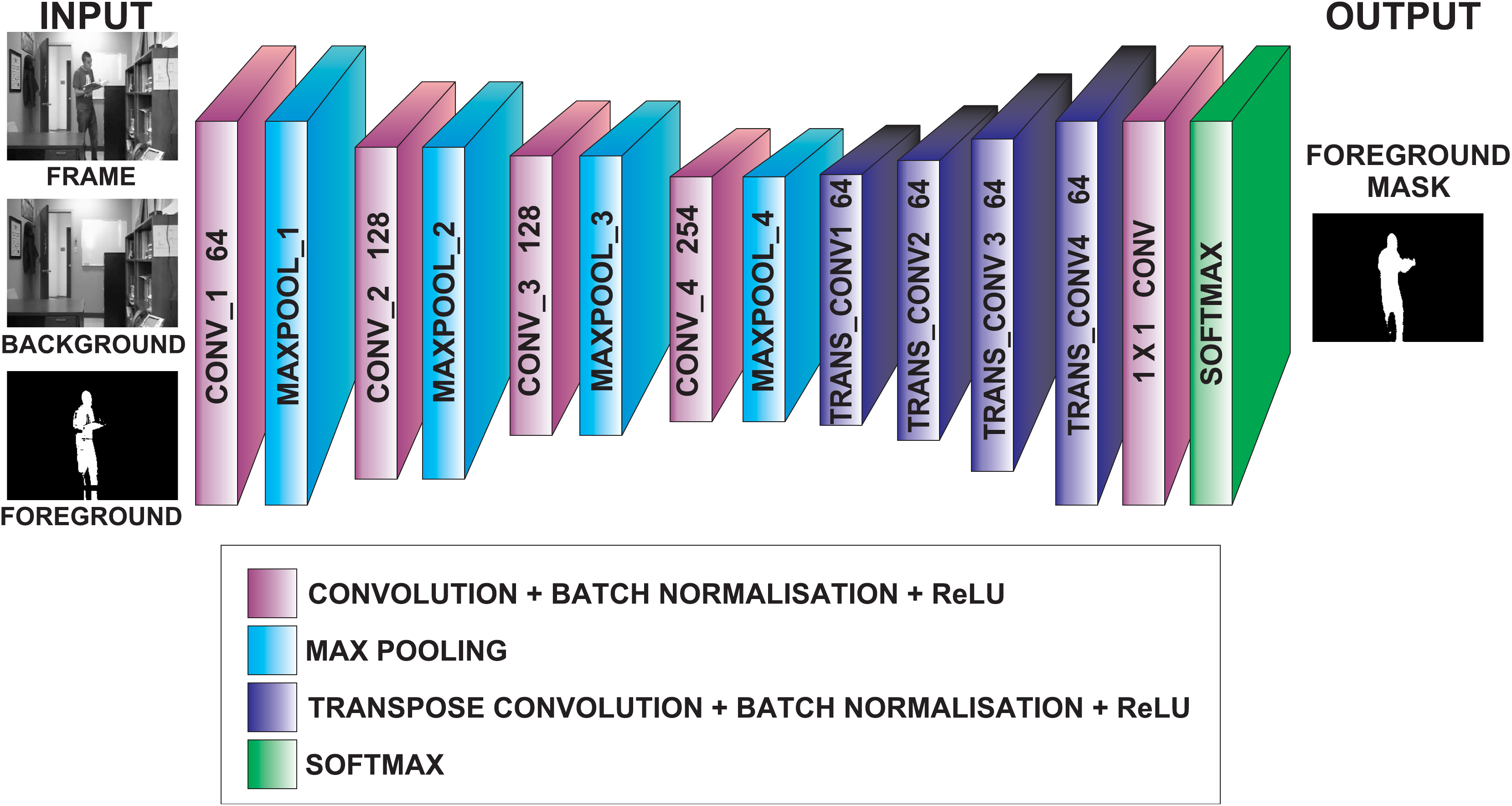}
\caption{Foreground segmentation network. The network receives three images concatenated in the channels dimension: input image, background image, and estimated foreground image. By concatenating an estimated foreground image to the input, we can run the model multiple times to iteratively improve the foreground estimation.}
\label{fig:net} %
\end{figure*}

To segment people or objects from the scene, we developed a background subtraction approach based on convolutional neural networks using MATLAB's Neural Network Toolbox. Figure \ref{fig:net} shows the architecture of the network. 

The network receives a three-channel input consisting of the input image, a background image, and an estimated foreground image. We build an encoder-decoder architecture trained from scratch and tackle the problem as a pixel-wise classification. By concatenating an estimated foreground image to the input, we can run the model multiple times to iteratively improve the foreground estimation.
The output of the network is a foreground segmentation mask that we use to segment people or objects from the scene.

%.....................................
% IMAGE: CNN model
%.....................................
%.....................................

\subsubsection{Training}
\begin{figure}[!ht]
\centering
\includegraphics[width=0.35\textwidth]{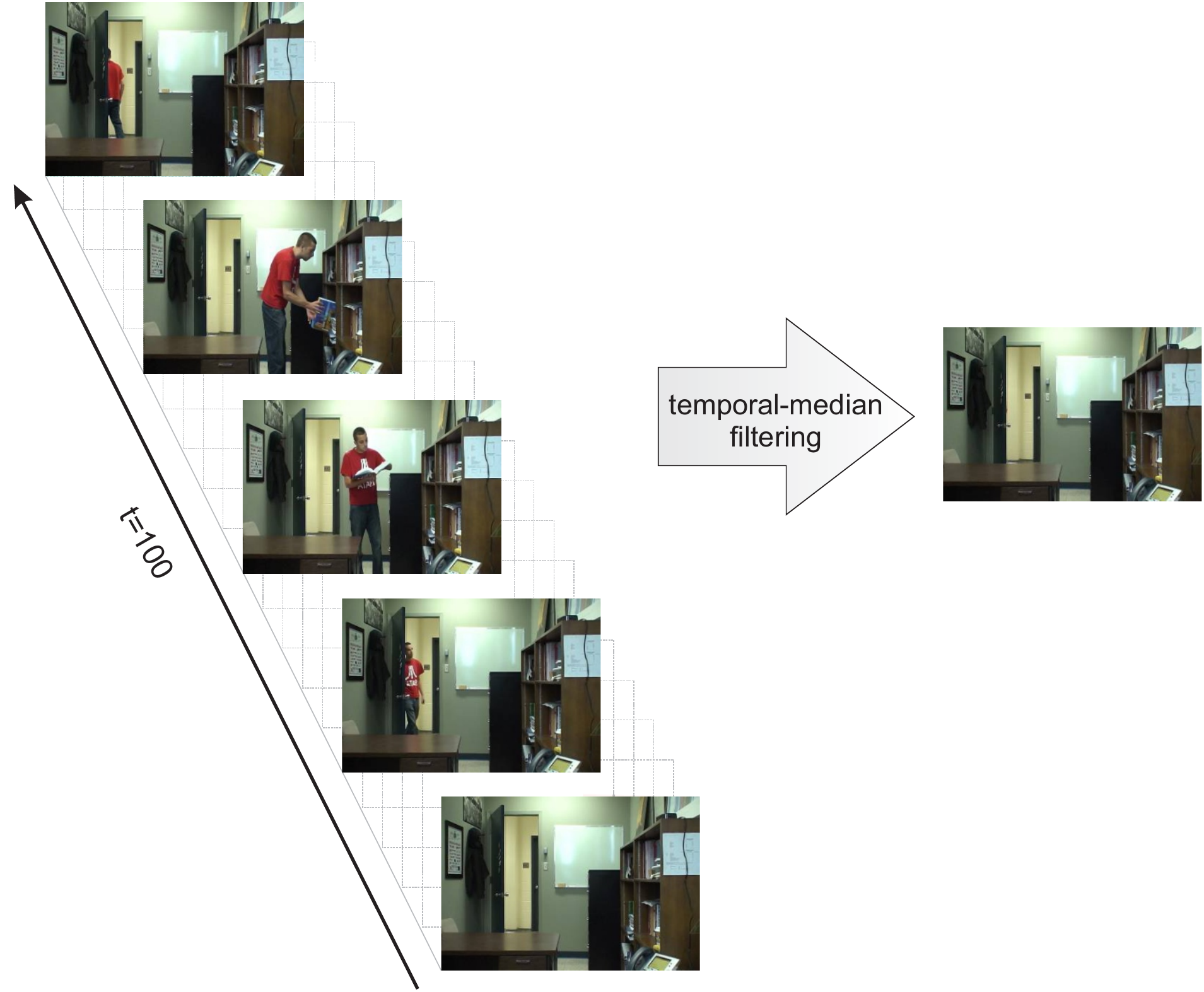}
\caption{Background image extraction using a temporal median filter. This filter keeps pixel intensities that appear in more than 50\% of the frames.}
\label{fig:median_filter}
\end{figure}

To train the network, we use the CDNet2014 dataset which consists of 54 videos divided into 11 video categories~\cite{goyette2012changedetection}.
Before training, we pre-compute the background images of each video using a temporal median filter where it is assumed that the background pixel intensities appear in more than 50\% of the input sequence. Figure~\ref{fig:median_filter} shows a schematic of this approach. 

%.....................................
% IMAGE: Median Filter
%.....................................
%.....................................

During training and testing, we generate the input foreground images by subtracting the background from the input image followed by a thresholding operation. For training, we create these foreground images with random morphological erosions or dilations to simulate imperfect foregrounds. 
We used data augmentation to provide more examples to the network. Concretely, we added random left or right reflection and random $x$ or $y$ translation of $\pm 10$ pixels.

Videos in CDnet2014 dataset have an imbalance between background and foreground. To handle this, we applied class balancing~\cite{eigen2015predicting,badrinarayanan2017segnet} in the cross-entropy loss (eq.~\ref{eq:loss})  where the weight assigned to a class in the loss function is the ratio of the
median of class frequencies computed on the entire training set divided by the class frequency.

%.....................................
% Loss function equation
%..................................... 
\begin{equation}\label{eq:loss}
loss(x, c) = w[c] \left(-x[c] + \log\left(\sum_j \exp(x[j])\right)\right),
\end{equation}
%..................................... 

where $c$ is the class (either background or foreground) and $w$ is the weight associated with the class.

We train the network for 100 epochs with an initial learning rate of $1e-4$ and a piecewise learning rate schedule with a drop factor of $0.5$ every $10$ epochs. We use a batch size of $12$ images and a $L_2$ regularization with a weight decay of $1e-4$. Training takes around three days on an NVIDIA Titan X GPU.

%%%%%%%%%%%%%%%%%%%%%%%%%%%%%%%%%%%%%%%%%%%%%%%%%%%%%%%%%%%%%%%%%%%%%%%%%%%%%
%%%%%%%%%%%%%%%%%%%%%%%%%%%%%%%%%%%%%%%%%%%%%%%%%%%%%%%%%%%%%%%%%%%%%%%%%%%%%
\subsection{Data Fusion and Simulated Holographic Display}
\label{sec:fusion}

The procedure for data fusion starts with the acquisition of a color and depth frames from each camera and the transmission of those frames to the main computer. Then, using our segmentation method, we obtain the foreground masks of the people or objects on the color images and map these masks to the depth images using Kinect SDK coordinate mapping (Kinect SDK provides a one-to-correspondence between color pixels and depth values and vice-versa). 

To obtain the 3D points, we calculate the $[x,y,z]$ coordinates of each pixel inside the foreground mask of the depth images using perspective projection with $z$ = depth. Then, we project the $[x,y,z]$ points onto the color frame using the intrinsic and extrinsic parameters of the color camera to extract the corresponding color of each 3D point. 

To fuse together the colored 3D data we use the extrinsic parameters of each camera, \emph{i.e.,} the poses between each camera and the reference, and use them to transform all the point clouds into a single reference frame.

\begin{figure}[!ht]
\centering
\includegraphics[width=0.25\textwidth]{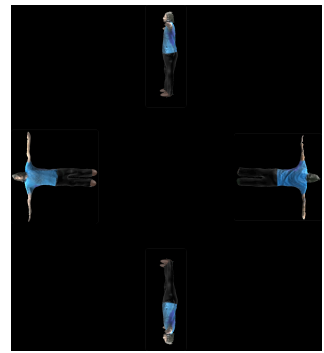}
\caption{Image projected on the pyramid. Each view is projected on each face of the pyramid.}
\label{fig:display2}       
\end{figure}

\begin{figure}[!ht]
\centering
\includegraphics[width=0.33\textwidth]{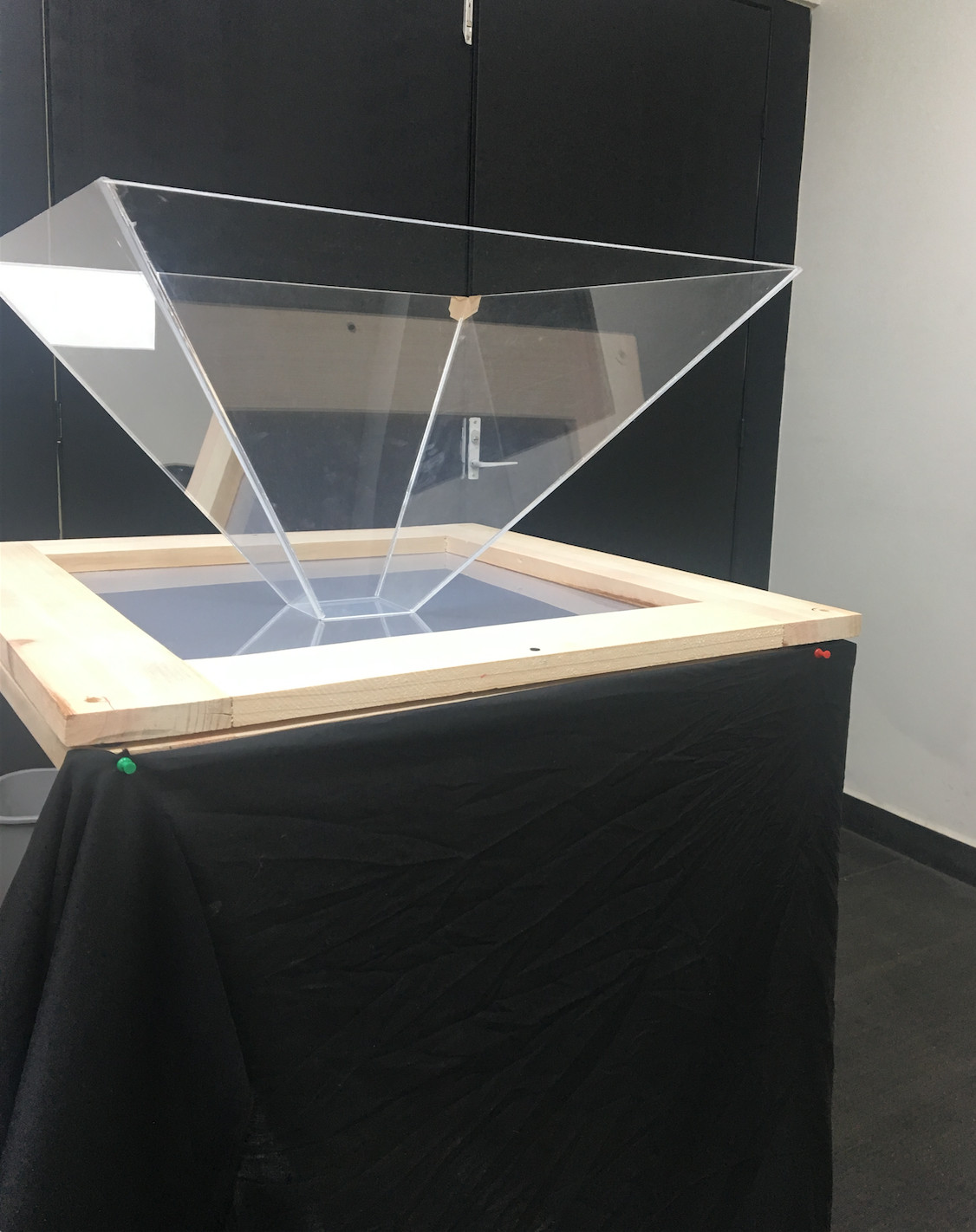}
\caption{Holographic display. Quadrangular pyramid made of acrylic with a polarized acrylic based. Below the pyramid is a video-projector that projects the video to the four faces of the pyramid.}
\label{fig:display1}
\end{figure}

Finally, to produce the simulated holographic visualization, we split the 3D model into four views (Figure \ref{fig:display2}) that are visualized in the corresponding face of the pyramid display.
Our display is a quadrangular pyramid made of acrylic that allows the creation of the hologram effect by projecting the images on an acrylic base using a video projector (see Figure \ref{fig:display1}). The display simulates the visualization of the 3D models from any angle.

%.....................................
% IMAGE: Projected images
%.....................................
%.....................................

%.....................................
% IMAGE: Pyramid
%.....................................
%.....................................

We built a wood structure that holds the acrylic base and the pyramid and contains the video projector inside. The acrylic base is polarized to reduce the light reflections generated by the video projector. 

We implemented all the system in MATLAB 2017b. For the data transmission, we used multi-threaded communication, for Kinect V2 data acquisition we use the Kin2 Toolbox~\cite{Terven2016kin2}, and for the foreground extraction and rendering, we use CPU-only. The whole system runs at 10 FPS.

%%%%%%%%%%%%%%%%%%%%%%%%%%%%%%%%%%%%%%%%%%%%%%%%%%%%%%%%%%%%%%%%%%%%%%%%%%%%%
%%%%%%%%%%%%%%%%%%%%%%%%%%%%%%%%%%%%%%%%%%%%%%%%%%%%%%%%%%%%%%%%%%%%%%%%%%%%%
\section{Results}
\label{sec:Results}
In this section, we describe the calibration results of the eight cameras (four RGB cameras + four depth cameras) that compose our system. Then, we provide quantitative results on the segmentation task followed by qualitative results of data fusion and the simulated holographic system. 

%%%%%%%%%%%%%%%%%%%%%%%%%%%%%%%%%%%%%%%%%%%%%%%%%%%%%%%%%%%%%%%%%%%%
\subsection{Calibration Results}
Our camera model includes intrinsic parameters (focal length $\alpha$, skew $\gamma$, and principal point $(u_0,v_0)$) and extrinsic parameters (rotation {\bf {R}} and translation {\bf {t}} with respect to the reference) plus three radial distortion parameters ($k_1$, $k_2$, $k_3$) and two tangential distortion parameters ($p_1$,$p_2$).
We obtain a mean re-projection error of $0.13$ pixels for the depth cameras and $0.46$ pixels for the color cameras.
Table \ref{table:calib_cam} shows the intrinsic parameters and distortion parameters for the color and depth cameras. 

%%%%%%%%%%%%%%%%%%%%%%%%%%%%%%%%%%
\begin{table*}[htb]
\caption{Calibration Results. Focal length ($\alpha$), skew ($\gamma$), principal point $(u_0,v_0)$, radial distortion parameters ($k_1$, $k_2$, $k_3$) and tangential distortion parameters ($p_1$,$p_2$).}
\centering
\scalebox{0.8}{
\begin{tabular}{lcccccccccc}
\toprule
Intrinsic parameters & {$\alpha$} & {$\gamma$} & {$u_0$} & {$v_0$} & {$k_1$} & {$k_2$} & {$k_3$} & {$p_1$} & {$p_2$} & {\it RMSE} \\ 
\midrule
\multicolumn{3}{l}{Color Cameras} & & & & & & & \\
\midrule
{\it Camera 1} & 1064.9131 & 0.0979 & 962.6340 & 537.3419 & 0.0145  & -0.0035  & 1.50e-04  & {-2.21e-06}  & {-2.04e-04}  & \bf{0.4627}\\
{\it Camera 2} & 1064.4064 & 0.3025 & 969.4298 & 556.0011 & {0.0198} & {-0.0444} &  4.81e-02 & {-3.52e-04}  &  {\phantom{+}1.25e-04} & \bf{0.4080}\\
{\it Camera 3} & 1060.5403 & 0.7103 & 966.5805 & 555.8031 & {0.0218} & {-0.0419} & {4.11e-02}  & {\phantom{+}1.75e-03} & {\phantom{+}1.82e-03} & \bf{0.4720}\\
{\it Camera 4} & 1064.1054 & 1.1388 & 962.1830 & 555.7498 & {0.0157} & {-0.2277} & {8.23e-03} & {-7.47e-04} & {-7.84e-04} & \bf{0.4931}\\
\midrule
\multicolumn{3}{l}{Depth Cameras} & & & & & & & \\
\midrule
{\it Camera 1} & 365.0738 & 0.0856 & 255.2444 & 215.5756 & 0.0854  & -0.2492  & 0.0783  & {1.41e-04}  & {-2.81e-04}  & \bf{0.1023}\\
{\it Camera 2} & 364.0859 & 0.0496 & 256.8059 & 215.3510 & {0.1158} & {-0.3668} &  0.2433 & {6.30e-04}  &  {-1.85e-04} & \bf{0.1080}\\
{\it Camera 3} & 363.2690 & 0.2245 & 258.4067 & 216.8210 & {0.1092} & {-0.3238} & {0.1494}  & {1.11e-04} & {\phantom{+}1.29e-04} & \bf{0.1236}\\
{\it Camera 4} & 365.5980 & 0.0189 & 254.9488 & 215.8214 & {0.0809} & {-0.2277} & {0.0511} & {-2.48e-04} & {-1.81e-04} & \bf{0.1974}\\
\bottomrule
\end{tabular}}
\label{table:calib_cam}
\end{table*}
%%%%%%%%%%%%%%%%%%%%%%%%%%%%%%%%%

To evaluate the performance of our calibration method, we reconstruct objects with known dimensions. For a cardboard box shown in Figure~\ref{fig:holograma}(a), we obtain a reconstruction error of $3$ mm in width and $2$mm in height.

\begin{figure*}[!t]
\centering
\subfloat[20 images]{\includegraphics[scale=0.5]{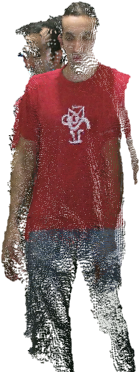}
\label{fig:im1}}
\hfil
\subfloat[50 images]{\includegraphics[scale=0.5]{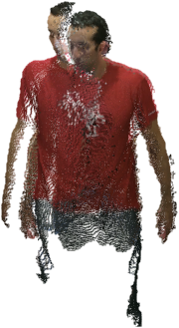}
\label{fig:im2}}
\hfil
\subfloat[70 images]{\includegraphics[scale=0.45]{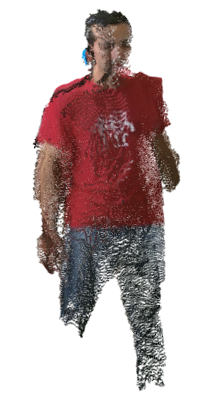}
\label{fig:im3}}
\hfil
\subfloat[100 images]{\includegraphics[scale=0.47]{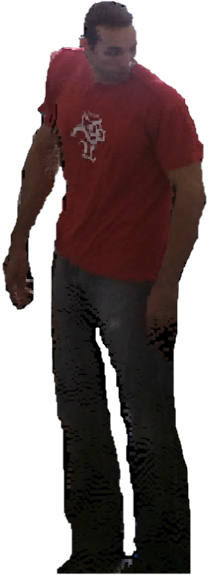}
\label{fig:ImJ}}
\caption{3D Avatar result when calibrating with (a) 20 images, (b) 50 images, (c) 70 images, and (d) 100 images.}
\label{fig:calib_pointclouds}
\end{figure*}

To provide a visually appealing 3D representation of the remote participant, we fuse the data from the multiple cameras. The quality of the fusion relies on the camera calibration results as shown in Figure~\ref{fig:calib_pointclouds}. \ref{fig:calib_pointclouds}(a) shows the fusion with calibration results using $20$ input images, \ref{fig:calib_pointclouds}(b) uses a calibration with $50$ images, \ref{fig:calib_pointclouds}(c) uses calibration with 70 images, and \ref{fig:calib_pointclouds}(d) uses 100 images. These results show that the extrinsic parameters tend to converge to the correct values with an increasing number of calibration images.

%.....................................
% FIGURE: Pointclouds for different calibrations
%.....................................
%.....................................

%%%%%%%%%%%%%%%%%%%%%%%%%%%%%%%%%%%%%%%%%%%%%%%%%%%%%%%%%%%%%%%%%%%%%%%%%%%%%
%%%%%%%%%%%%%%%%%%%%%%%%%%%%%%%%%%%%%%%%%%%%%%%%%%%%%%%%%%%%%%%%%%%%%%%%%%%%%
\subsection{Foreground Extraction Results}
To evaluate the performance of our foreground extraction method, we calculate the following metrics: recall, specificity, false positive rate (FPR), percentage of bad classifications (PBC), precision, and f-score ($F_1$). Table \ref{table:performance_cdnet} shows these metrics for 9 categories of CDNet2014 dataset. We omit results from the \emph{PTZ} and \emph{Camera Jitter} categories since our method assumes the cameras are static.
All these metrics are based on the correctly/incorrectly classified pixels which can be defined with the true positives (TP), false positives (FP), true negatives (TN), and false negatives (FN) described below~\cite{babaee2017deep,sakkos2017end}:

\begin{itemize}
\item True Positives (TP) consist of the foreground pixels in the output segmentation that are also foreground pixels in the ground truth segmentation.
\item False Positives (FP) consist of the foreground pixels in the output segmentation that are not foreground pixels in the ground truth segmentation.
\item True Negatives (TN) consist of the background pixels in the output segmentation that are also background pixels in the ground truth segmentation.
\item False Negatives (FN) consist of the background pixels in the output segmentation that are not background pixels in the ground truth segmentation.
\end{itemize}

The equations for the metrics shown in Table \ref{table:performance_cdnet} are the following:

% Recall
\begin{ceqn}
\begin{align}
Recall=\frac{TP}{TP + FN}
\end{align}
\end{ceqn}

% Specificity
\begin{ceqn}
\begin{align}
Specificity=\frac{TN}{TN + FP}
\end{align}
\end{ceqn}

% FPR
\begin{ceqn}
\begin{align}
FPR=\frac{FP}{FP + TN}
\end{align}
\end{ceqn}

% PBC
\begin{ceqn}
\begin{align}
PBC=100 \times \frac{FN +FP}{TP + FN + FP + TN}
\end{align}
\end{ceqn}

% Precision
\begin{ceqn}
\begin{align}
Precision=\frac{TP}{TP + FP}
\end{align}
\end{ceqn}

% F1 score
\begin{ceqn}
\begin{align}
F_{1}=\frac{Precision \times Recall}{Precision + Recall},
\end{align}
\end{ceqn}
%..................................... 

Table \ref{table:performance_cdnet_other_methods} compares the performance of our method against other methods of the CDNet dataset in 9 categories namely baseline (BSL), dynamic background (DBG), intermittent object motion (IOM), shadow (SHD), thermal (THM), bad weather (BDW), low frame-rate (LFR), night videos (NVD), and turbulence (TBL).
Although we did not achieve state-of-the-art results on the general background subtraction task, our results are sufficient for our purposes, and the model can run in real-time due to the small network size. 

Our background subtraction method is freely available in \footnote{\url{github.com/jrterven/backsub}}.
%.....................................
% TABLE: CDNet 2014. Per class accuracy
%.....................................
\begin{table*}[htb]
\caption{Segmentation Results on CDNet2014}
\centering
\scalebox{0.9}{
\begin{tabular}{lccccccc}
\toprule
Category & Recall & Specificity & FPR & FNR & PBC & Precision & $F_1$ \\ 
\midrule
{\it Baseline} & 0.9391 & 0.9989 & 0.0010 & 0.0609 & 0.2126 & 0.9681 & 0.9530
 \\
%{\it Camera Jitter} & 0.9844 & 0.8770 & 0.1230
%& 0.0155 & 11.7690 & 0.2946 & 0.4423 \\
{\it Dynamic background} & 0.7633 & 0.9947 & 0.0052 & 0.2366 & 0.6759 & 0.7214 & 0.7076 \\
{\it Intermittent object motion} & 0.7584 & 0.9843 & 0.0156 & 0.2416 & 3.2235 & 0.7578 & 0.7040 \\
{\it Shadow} & 0.9815 & 0.9916 & 0.0083 & 0.0184 & 0.8903 & 0.8228 & 0.8919 \\
{\it Thermal} & 0.7796 & 0.9963 & 0.0036 & 0.2204 & 1.1554
& 0.9288 & 0.8338\\
{\it Bad weather} & 0.9264 & 0.9968 & 0.0032 & 0.0736 & 0.4188 & 0.8411 & 0.8781 \\
{\it Low frame-rate} & 0.8966 & 0.9801 & 0.0199 & 0.1033 & 2.1999 & 0.6987 & 0.6961 \\
{\it Night } & 0.8804 & 0.9897 & 0.0103 & 0.1196 & 1.3152 & 0.6622 & 0.7401 \\
{\it Turbulence} & 0.9845 & 0.9911 &  0.0088 & 0.0155 & 0.8967 & 0.4415 & 0.5752 \\
{\it Overall} & 0.8896 & 0.9802 & 0.0198 & 0.1104  & 2.2600 & 0.7153 & 0.7432 \\
\bottomrule
\end{tabular}}
\label{table:performance_cdnet}
\end{table*}
%.....................................

%.....................................
% TABLE: Comparing Background subtraction accuracy with other methods
%.....................................
\begin{table*}[htb]
\caption{$F_1$ score comparison of different background subtraction methods over 9 categories of CDNet2014: baseline(BSL), dynamic background(DBG), intermittent object motion(IOM), shadow(SHD), thermal(THM), bad weather(BDW), low frame-rate(LFR), night videos(NVD), and turbulence(TBL).}
\centering
\scalebox{0.9}{
\begin{tabular}{lccccccccc}
\toprule
Method & $F_{BSL}$ & $F_{DBG}$  & $F_{IOM}$ & $F_{SHD}$ & $F_{THM}$ & $ F_{BDW}$ & $F_{LFR}$ & $F_{NVD}$  & $F_{TBL}$ \\ 
\midrule
{\it OURS} & 0.9530  &  0.7076  &  0.7040 &  0.8919  & 0.8338  & 0.8781 & 0.6961 & 0.7401  & 0.5752   \\
{\it 3D} \cite{sakkos2017end} & 0.9509  &  \bf{0.9614}  &  \bf{0.9698} &  \bf{0.9706}  & \bf{0.9830}  & \bf{0.9509} & \bf{0.8862} & \bf{0.8565}  & \bf{0.8823}   \\
{\it CNN} \cite{babaee2017deep}  & \bf{0.9580} & 0.8761  & 0.6098 & 0.9304 & 0.7583 & 0.8301 & 0.6002 & 0.5835  & \bf{0.8455} \\
{\it PBAS} \cite{hofmann2012background}  & 0.9242 & 0.6829  & 0.5745 & 0.8143 & 0.7556 & 0.7673 & 0.5914 & 0.4387  & 0.6349 \\
{\it PAWCS} \cite{Charles2015} & 0.9397 & 0.8938 & 0.7764 & 0.8913 & 0.8324  & 0.8152 & 0.6588 & 0.4152  & 0.6450\\
{\it SuBSENSE} \cite{Charles2015SuBSENSE}  & 0.9503 & 0.8177  & 0.6569 & 0.8986 & 0.8171 &  0.8619 & 0.6445 & 0.5599  & 0.7792\\
{\it MBS} \cite{Sajid2015} & 0.9287 & 0.7915  & 0.7568 & 0.8262 & 0.8194  & 0.7730 & 0.6279 & 0.5158  & 0.5698\\
{\it GMM} \cite{stauffer1999adaptive} & 0.8245 & 0.6330  & 0.5207 & 0.7156 & 0.6621  & 0.7380 & 0.5373 & 0.4097 & 0.4663 \\
{\it RMoG } \cite{VARADARAJAN20153488} & 0.7848 & 0.7352  & 0.5431 & 0.7212 & 0.4788 &0.6826 & 0.5312 & 0.4265 & 0.4578 \\
{\it Spectral-360} \cite{sedky2010image} & 0.9330 & 0.7872  & 0.5656 & 0.8843 & 0.7764 & 0.7569 & 0.6437 & 0.4832  & 0.5429 \\
\bottomrule

\end{tabular}}
\label{table:performance_cdnet_other_methods}
\end{table*}
%.....................................

%%%%%%%%%%%%%%%%%%%%%%%%%%%%%%%%%%%%%%%%%%%%%%%%%%%%%%%%%%%%%%%%%%%%%%%%%%%%%
%%%%%%%%%%%%%%%%%%%%%%%%%%%%%%%%%%%%%%%%%%%%%%%%%%%%%%%%%%%%%%%%%%%%%%%%%%%%%
\subsection{Qualitative Results}
\begin{figure*}[!t]
\centering
\subfloat[Test object 1: cardboard box]{\includegraphics[width=1.5in]{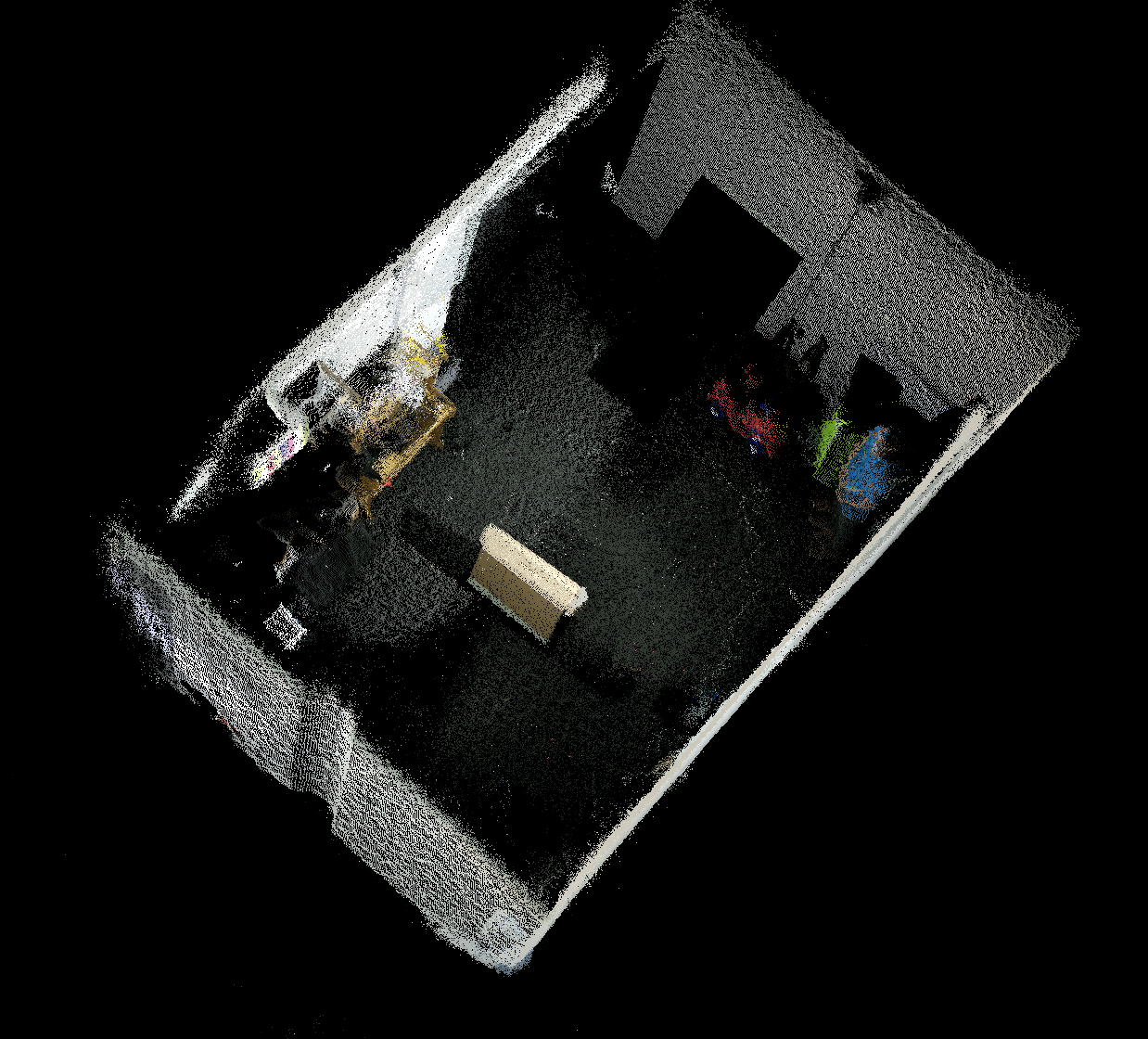}
\label{fig:pc1}}
\hfil
\subfloat[Test object 2: a person]{\includegraphics[width=1.4in]{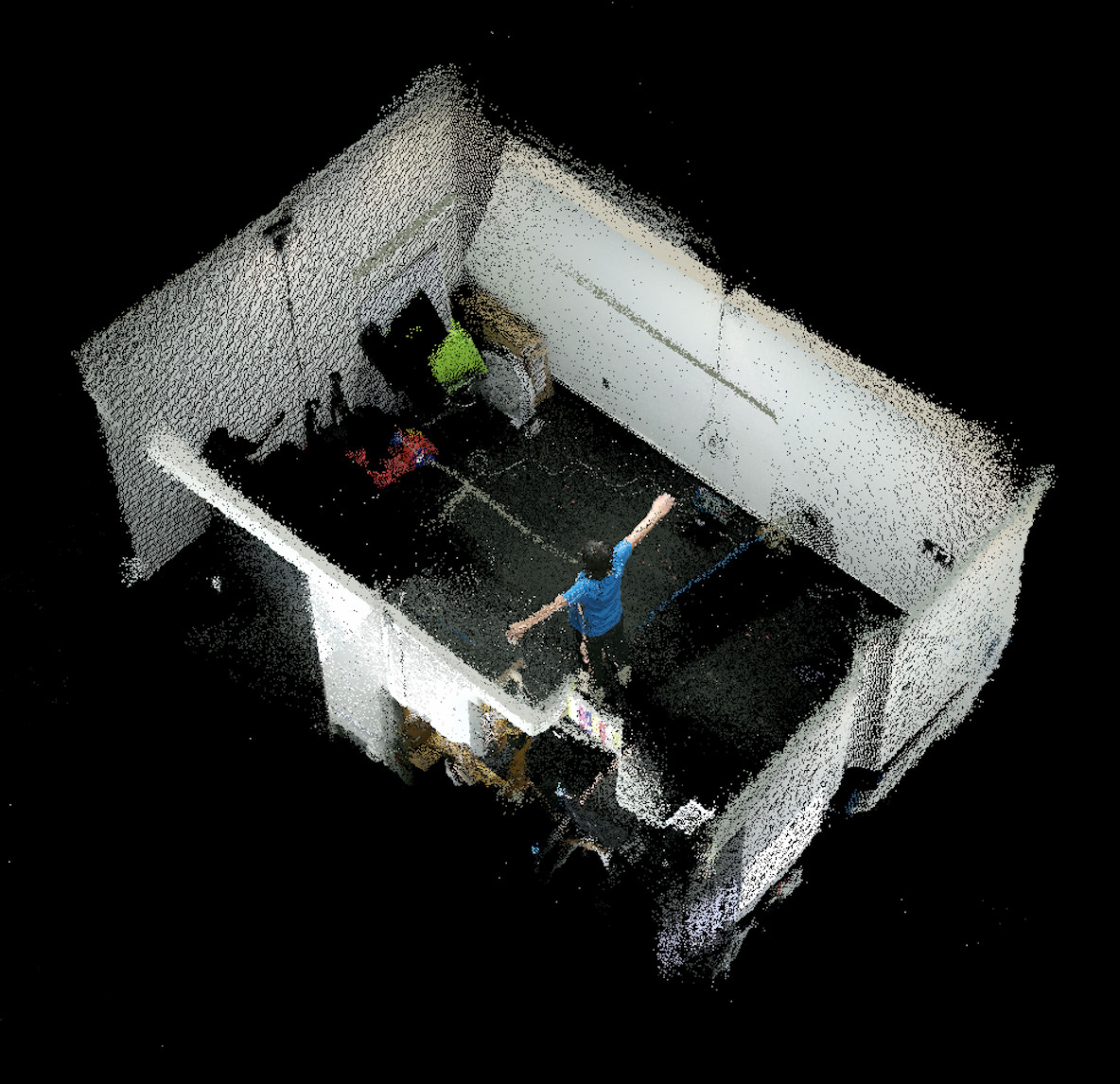}
\label{fig:pc2}}
\hfil
\subfloat[Test object 3: floor fan]{\includegraphics[width=1.5in]{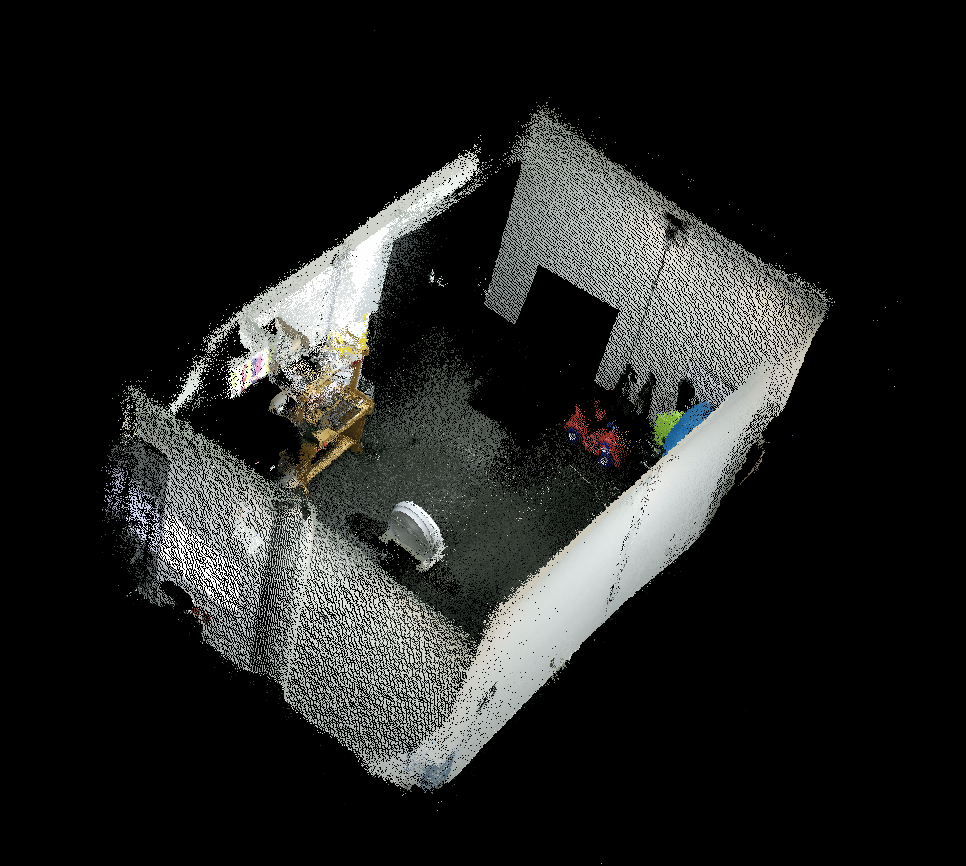}
\label{fig:pc3}}
\caption{Point clouds of different test objects obtained with data fusion from the four RGB-D cameras.}
\label{fig:pointclouds}
\end{figure*}

Figure \ref{fig:pointclouds} shows qualitative results of 3D data fusion without foreground extraction. This figure shows a correct alignment of the eight cameras for different objects in the scene. 
%Figure \ref{fig:pc1} shows the results for a cardboard box, Figure \ref{fig:pc2} shows a person inside the environment, and Figure \ref{fig:pc3} shows a floor fan. 

%.....................................
% FIGURE: Qualitative results
%.....................................
%.....................................

\begin{figure*}[!t]
\centering
\subfloat[Cardboard box]{\includegraphics[width=1.5in]{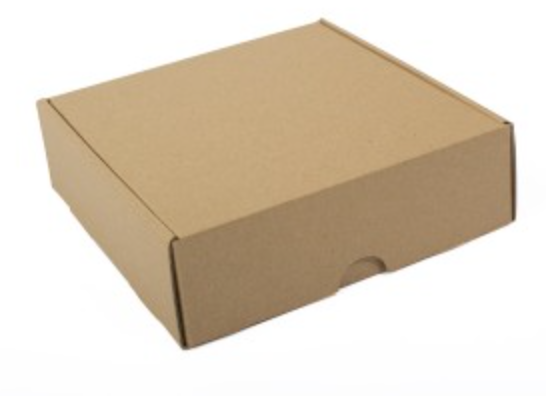}
\label{fig:pch1}}
\hfil
\subfloat[segmentation and reconstruction]{\includegraphics[width=1.2in]{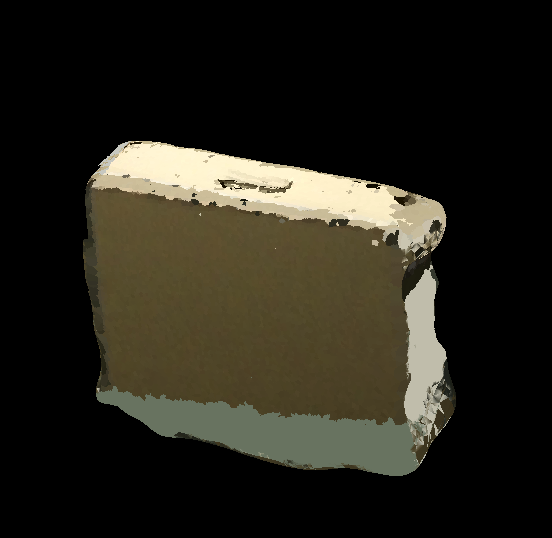}
\label{fig:pch2}}
\hfil
\subfloat[Holographic visualization]{\includegraphics[width=1.5in]{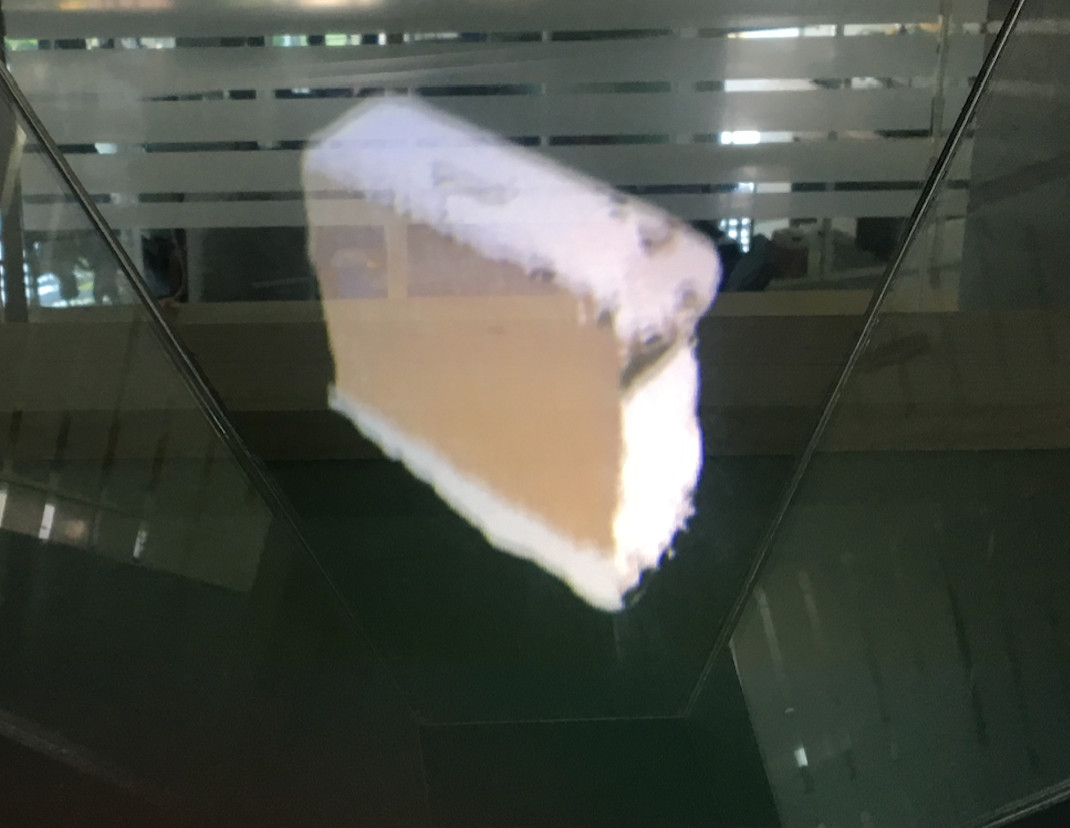}
\label{fig:pch3}}
\caption{The holographic visualization of the fused 3D models. (a) shows the real object, (b) shows its fusion and 3D reconstruction, and (c) shows the final visualization projected in the pyramid.}
\label{fig:holograma}
\end{figure*}

\begin{figure*}[!t]
\centering
\subfloat[Floor fan]{\includegraphics[width=1in]{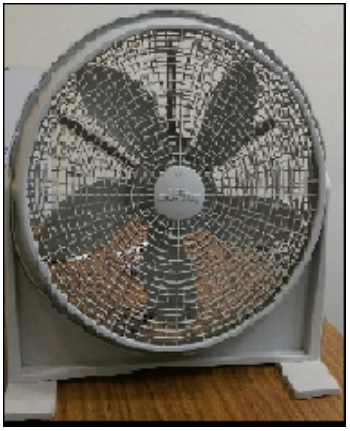}
\label{fig:ph1}}
\hfil
\subfloat[3D reconstruction]{\includegraphics[width=1.2in]{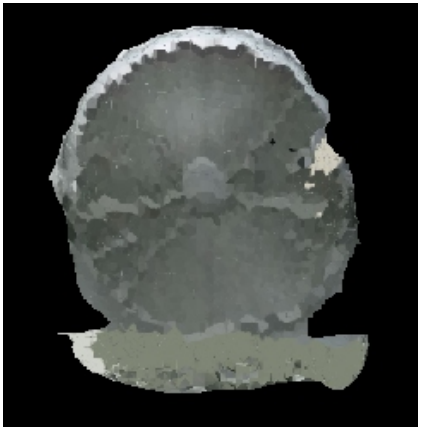}
\label{fig:ph2}}
\hfil
\subfloat[Holographic visualization]{\includegraphics[width=1.2in]{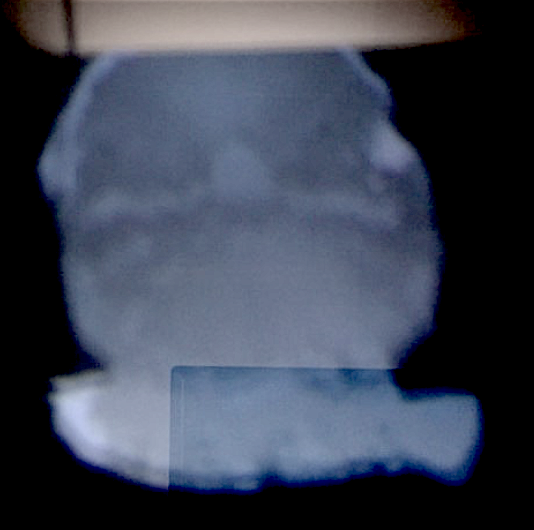}
\label{fig:ph3}}
\caption{The holographic visualization of the fused 3D models. (a) shows the real object, (b) shows its fusion and 3D reconstruction, and (c) shows the final visualization projected in the pyramid.}
\label{fig:holograma2}
\end{figure*}

The simulated holographic display generates a  $360^{\circ}$ visualization of the fused 3D models. Figures \ref{fig:holograma} and \ref{fig:holograma2} shows examples of holographic visualizations of real objects. (a) shows the real object, (b) shows the reconstruction obtained from fusing the 3D reconstructions of the segmented objects, and (c) shows the final visualization projected in the holographic display.

%.....................................
% FIGURE: Holograms
%.....................................
%.....................................

%.....................................
% FIGURE Holographic objects
%.....................................
%.....................................

\begin{figure}[!ht]
\centering
\includegraphics[width=0.33\textwidth]{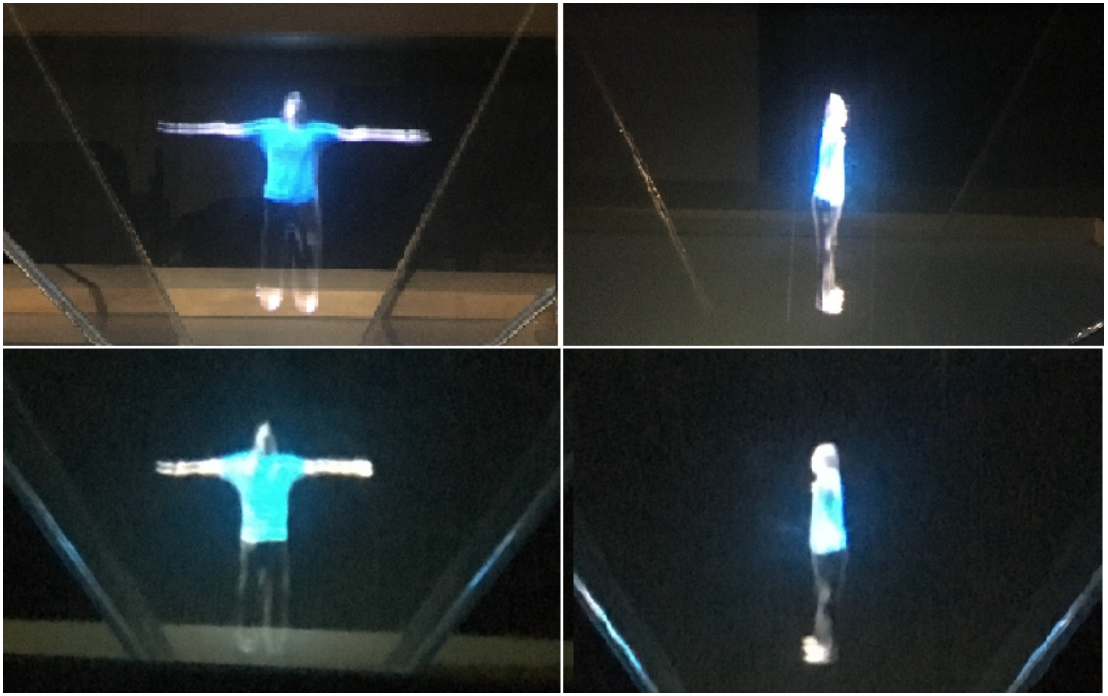}
\caption{Holographic display shows the four views of the holographic pyramid displaying the 3D reconstruction of the person.}
\label{fig:reco_holo}
\end{figure}

Figure \ref{fig:reco_holo} shows the four views of the holographic pyramid displaying the 3D reconstruction of a person.

%.....................................
% IMAGE: Holgraphic person
%.....................................
%.....................................

%%%%%%%%%%%%%%%%%%%%%%%%%%%%%%%%%%%%%%%%%%%%%%%%%%%%%%%%%%%%%%%%%%%%%%%%%%%%%
%%%%%%%%%%%%%%%%%%%%%%%%%%%%%%%%%%%%%%%%%%%%%%%%%%%%%%%%%%%%%%%%%%%%%%%%%%%%%
\section{Conclusions}
We presented a telepresence system that enables remote participants, represented as full 3D models projected into a custom-made simulated holographic display, to engage in real-time, co-present interaction. Our system does not require participants to wear any specialized equipment such as head mounted displays, enabling them to
move freely and view each other from different angles.

Our system is currently limited to 3D point cloud visual-only information. Our goal is to extend the system to render temporal consistent 3D meshes coupled with high fidelity audio transmission. 

We believe that our approach is an essential step towards democratizing high-fidelity full 3D telepresence because rather than relying on expensive technologies such as real holographic displays or high-end hardware, we pursue an approach that uses off-the-shelf components.

\section{acknowledgements}
This work was supported by CONACYT through postdoctoral support number 291113. We also want to thank CIDESI for providing the facilities and assistance during the development of this project.

\bibliographystyle{ieeetr}
\bibliography{biblio}

\end{document}